\documentclass{article} 
\usepackage{iclr2022_conference,times}

\usepackage{amsmath,amsfonts,bm}

\def\eqref#1{equation~\ref{#1}}

\def\1{\bm{1}}

\DeclareMathAlphabet{\mathsfit}{\encodingdefault}{\sfdefault}{m}{sl}
\SetMathAlphabet{\mathsfit}{bold}{\encodingdefault}{\sfdefault}{bx}{n}

\usepackage{hyperref}
\usepackage{url}
\usepackage{caption}
\usepackage{subcaption}

\usepackage{booktabs}
\usepackage{babel,adjustbox,booktabs,multirow}
\usepackage{wrapfig, lipsum}
\usepackage{comment}

\usepackage{multirow}
\usepackage{siunitx}

\usepackage{listings}

\title{Cascaded Fast and Slow Models for Efficient Semantic Code Search}

\author{Akhilesh Deepak Gotmare, Junnan Li, Shafiq Joty \& Steven C.H. Hoi 
\\
Salesforce Research Asia\\
\texttt{\{akhilesh.gotmare,junnan.li,sjoty,shoi\}@salesforce.com}
}

\newcommand{\shafiq}[1]{\textcolor{magenta}{#1}}

\iclrfinalcopy 
\begin{document}

\maketitle

\begin{abstract}


The goal of natural language semantic code search is to retrieve a semantically relevant code snippet from a fixed set of candidates using a natural language query. Existing approaches are neither effective nor efficient enough towards a practical semantic code search system. In this paper, we propose an efficient and accurate semantic code search framework with cascaded fast and slow models, in which a fast transformer encoder model is learned to optimize a scalable index for fast retrieval followed by learning a slow classification-based re-ranking model to improve the performance of the top K results from the fast retrieval. To further reduce the high memory cost of deploying two separate models in practice, we propose to jointly train the fast and slow model based on a single transformer encoder with shared parameters. The proposed cascaded approach is not only efficient and scalable, but also achieves state-of-the-art results with an average mean reciprocal ranking (MRR) score of $0.7795$ (across 6 programming languages) as opposed to the previous state-of-the-art result of $0.713$ MRR on the CodeSearchNet benchmark.  

\end{abstract}

\section{Introduction}

Building tools that enhance software developer productivity has recently garnered a lot of attention in the deep learning research community. Parallel to the progress in natural language processing, pre-trained language models (LM) like CodeBERT \citep{feng2020codebert}, CodeGPT \citep{lu2021codexglue}, CodeX \citep{chen2021evaluating}, PLBART \citep{ahmad-etal-2021-unified} and CodeT5 \citep{wang2021codet5} have now been proposed for understanding and generation tasks involving programming languages. 

Recent work on code generation like \cite{chen2021evaluating}'s 12B parameter CodeX and \cite{austin2021program}'s 137B parameter LM use large scale autoregressive language models to demonstrate impressive capabilities of generating multiple lines of code from natural language descriptions, well beyond what previous generation models like GPT-C \citep{svyatkovskiy2020intellicode} could accomplish. However, this impressive performance is often predicated on being able to draw many samples from the model and machine-check them for correctness. This setup will often not be the case in practice. Code generation models also entail security implications (possibility of producing vulnerable or misaligned code) making their adoption tricky. 

Given this current landscape, code retrieval systems can serve as attractive alternatives when building tools to assist developers. With efficient implementations, code search for a single query can typically be much faster for most practical index sizes than generating code with large scale LMs. As opposed to code generation, code retrieval offers the possibility of a much greater control over the quality of the result - the index entries can be verified beforehand. Leveraging additional data post training is easier when working with code search systems as this would simply require extending the index by encoding the new instances. Code search systems can be particularly of value for organizations with internal proprietary code. Indexing source code data internally for search can prevent redundancy and boost programmer productivity. A recent study by \cite{xu2021ide} surveys developers to understand the effectiveness of code generation and code retrieval systems. Their results indicate that the two systems serve complementary roles and developers prefer retrieval modules over generation when working with complex functionalities, thus advocating the need for better code search systems.


\begin{figure}[t!]
\centering
\includegraphics[width=0.9\textwidth]{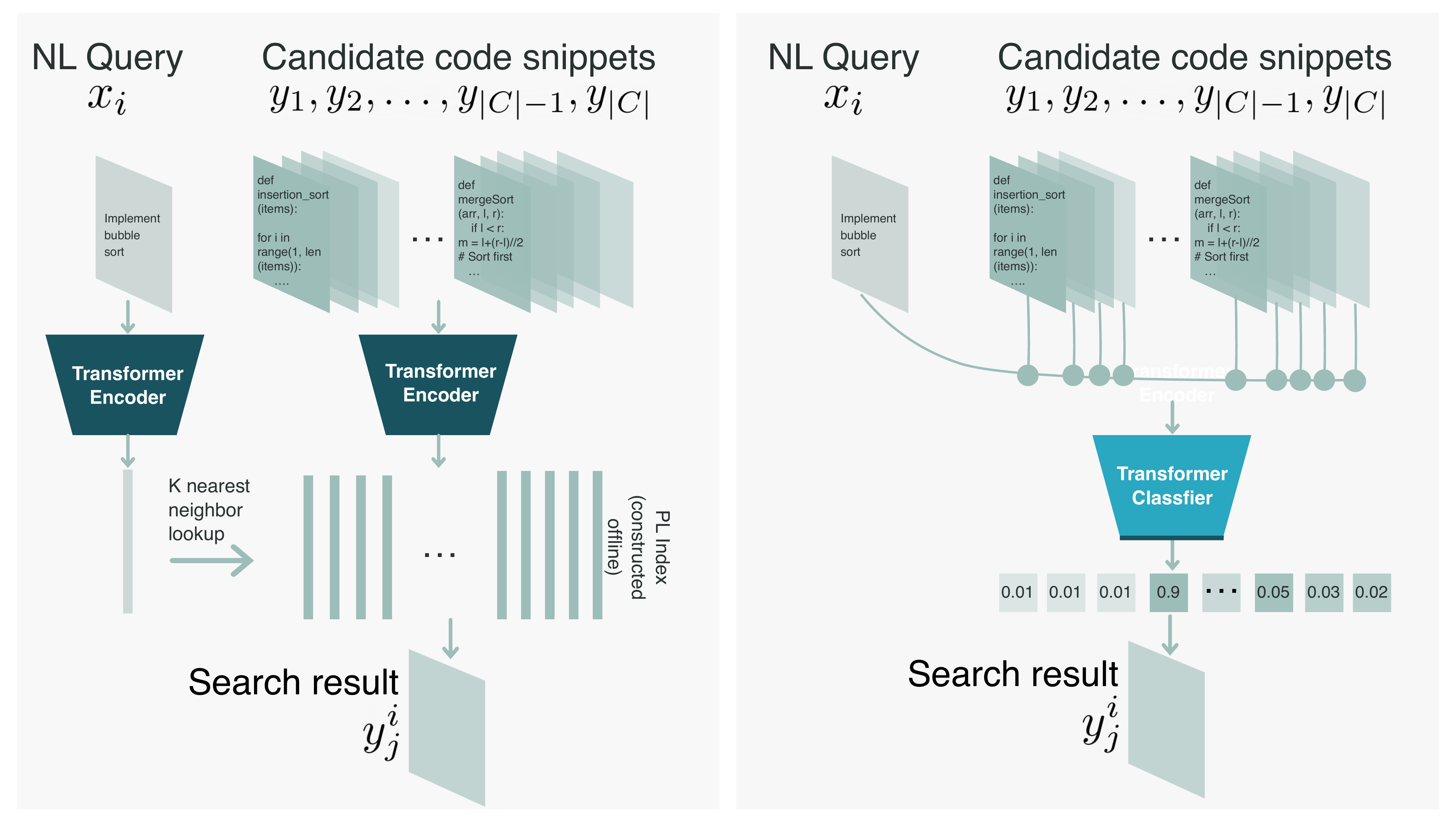} 
\caption{Illustration of the \textit{fast encoder} (left) and \textit{slow classifier} (right) based semantic code search approaches (at inference stage). With the encoder based approach, we independently compute representations of the NL query and candidate code sequences. The code snippet with representation nearest to the query vector is then returned as the search result. With the classifier based approach, we jointly process the query with each code sequence to predict the probability of the code matching the query description. The code sequence corresponding to the highest classifier confidence score is then returned as the search result.}
\label{fig_fast_slow}
\end{figure}

\begin{wrapfigure}{r}{0.55\textwidth}
    \centering
 \includegraphics[width=0.5\textwidth]{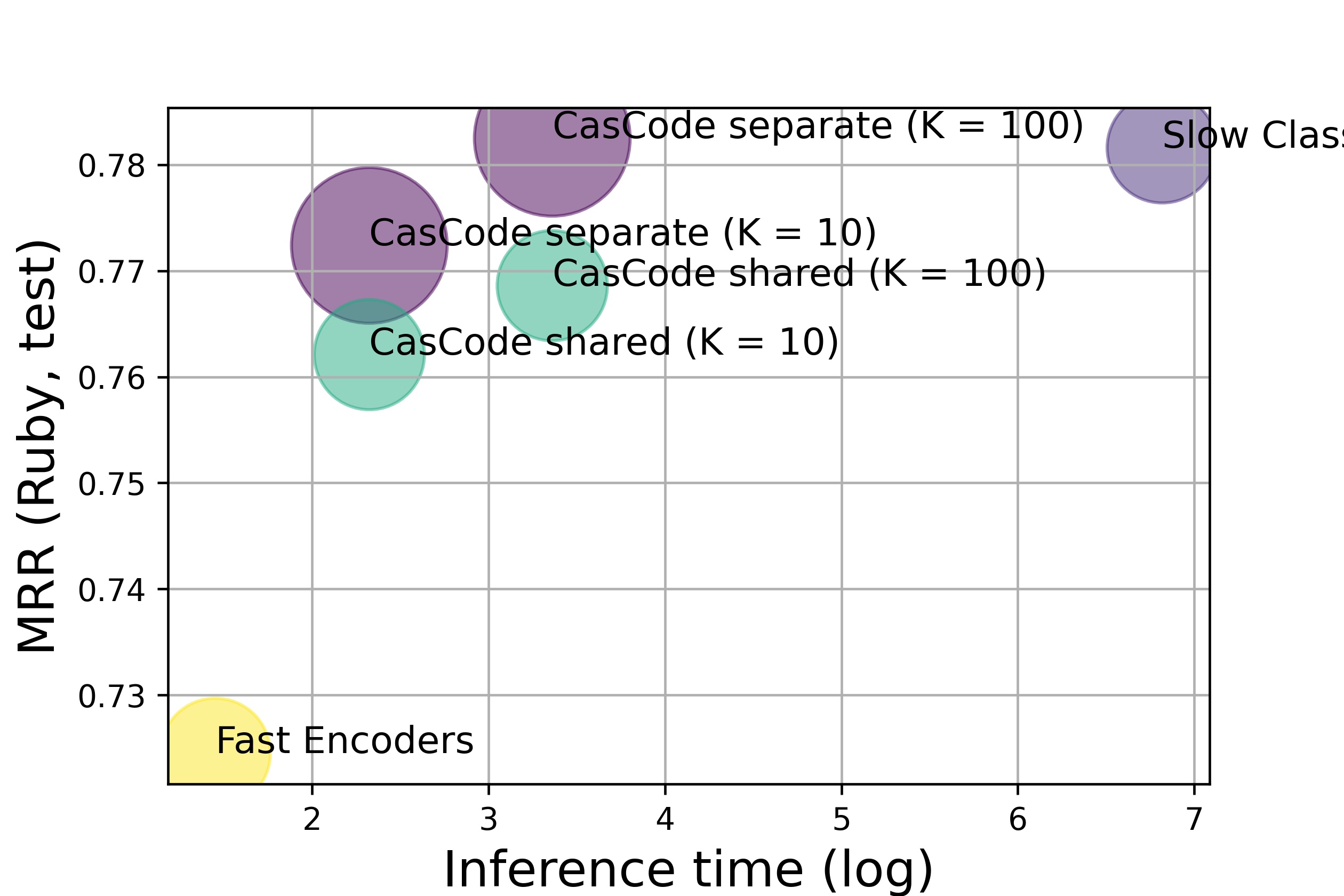}
\caption{Overview of the speed versus performance trade-off of current code search approaches. Areas of the circles here are proportional to model sizes. With CasCode, we are able to achieve performance comparable to the optimal classifier based approach (top right), while requiring substantially lesser inference time.}
\label{fig_fast_slow_tradeoff}
\end{wrapfigure}

Neural approaches to code search \citep{sachdev2018retrieval, guo2020graphcodebert, ye2016word, gu2018deep} involve encoding query and code independently into dense representations in the same semantic space. Retrieval is then performed using representational similarity (based on cosine or euclidean distances) of these dense vectors. An orthogonal approach involves encoding the query and the code jointly and training semantic code search systems as binary classifiers that predict whether a code answers a given query \citep{lu2021codexglue, huang2021cosqa}. With this approach, the model processes the query paired with each candidate code sequence. Intuitively, this approach helps in sharpening the cross information between query and code and is a better alternative for capturing matching relationships between the two modalities (natural language and programming language) than simple similarity metric between the encoder based sequence representations. 

While this latter approach can be promising for code retrieval,
previous methods have mostly leveraged it for binary classification tasks involving NL-PL sequence pairs. Directly adapting this approach to code search tasks would be impractical due to the large number of candidates to be considered for each query. We depict the complementary nature of these approaches in Figure \ref{fig_fast_slow} when using a transformer \citep{VaswaniSPUJGKP17} encoder based model for retrieval and classification. 

In order to leverage the potential of such nuanced classifier models for the task of retrieval, we propose a cascaded scheme (CasCode) where we process a limited number of candidates with the classifier model. This limiting is performed by employing the encoder based approach and picking its top few candidate choices from the retrieval set for processing by the second classifier stage. Our cascaded approach leads to state of the art performance on the CodeSearchNet benchmark with an overall mean reciprocal ranking (MRR) score of $0.7795$, substantially surpassing previous results. We propose a variant of the cascaded scheme with shared parameters, where a single transformer model can serve in both the modes - encoding and classification. This shared variant substantially reduces the memory requirements, while offering comparable retrieval performance with an MRR score of $0.7700$. 

Figure \ref{fig_fast_slow_tradeoff} illustrates the trade off involved between inference speed and MRR for different algorithmic choices, where we have the (\textit{fast}) encoder model on one extreme, and the (\textit{slow}) classifier model on the other. With CasCode, we offer performance comparable to the optimal scores attained by the classifier model, while requiring substantially lesser inference time, thus making it computationally feasible. Our codebase will be made publicly available for research purposes.

\section{Background}
\label{background_and_related_work}

Early work on neural approaches to code search include \cite{sachdev2018retrieval} who used unsupervised word embeddings to construct representations for documents (code snippets), followed by \cite{cambronero2019deep}'s supervised approach leveraging the pairing of code and queries. \cite{feng2020codebert} proposed pre-training BERT-style \citep{devlin-etal-2019-bert} masked language models with unlabeled (and unpaired) source code and docstrings, and fine-tuning them for text-to-code retrieval task. With this approach, the query representation can be compared during inference against a pre-constructed index of code representations and the nearest instance is returned as the search result. \cite{miech2021thinking} and \cite{ALBEF} have previously proposed similar approaches for text-to-visual retrieval. 

\cite{guo2020graphcodebert} leverage pairs of natural language and source code sequences to train text-to-code retrieval models. They adopt the contrastive learning framework \citep{chen2020simple} to train the retrieval model, where representations of natural language (NL) and programming language (PL) sequences that match in semantics (a positive pair from the bimodal dataset) are pulled together, while representations of negative pairs (randomly paired NL and PL sequences) are pushed apart. The infoNCE loss (a form of contrastive loss function \citep{gutmann2010noise}) used for this approach can be defined as follows:
\begin{align}
\label{eqn:contrastive}
\mathcal{L}_{\text{infoNCE}} = \frac{1}{N} \sum_{i=1}^{N} -\log \frac{\exp (f_{\theta}(x_i)^T f_{\theta}(y_i)/ \sigma)}{\sum_{j \in B} \exp(f_{\theta}(x_i)^T f_{\theta}(y_j)/\sigma)}
\end{align}
where $f_{\theta}(x_i)$ is the dense representation for the NL input $x_i$, and $y_i$ is the corresponding semantically equivalent PL sequence. $N$ is the number of training examples in the bimodal dataset, $\sigma$ is a temperature hyper-parameter, and $B$ denotes the current training minibatch. 

While the above approach applies for any model architecture, \cite{guo2020graphcodebert} employ GraphCodeBERT (a structure-aware transformer encoder pre-trained on code) and CodeBERT for $f_{\theta}$ in their experiments. We refer to this approach as the one using \textit{fast encoders} for retrieval. During inference, we are given a set of candidate code snippets $\mathcal{C} = \{y_1, y_2, \dots y_{|\mathcal{C}|} \}$, which are encoded offline into an index $\{ f_{\theta}(y_j)~~\forall j \in \mathcal{C}\}$. For a test NL query $x_i$, we then compute $f_{\theta}(x_i)$ and return the code snippet from $\mathcal{C}$ corresponding to the nearest neighbor (as per some distance metric e.g. cosine similarity) in the index. The rank $r_i$ assigned to the correct code snippet (for the query $x_i$) from $\mathcal{C}$ is then used to compute the mean reciprocal ranking (MRR) metric $\frac{1}{N_{test}}\sum_{i=1}^{N_{test}} \frac{1}{r_i}$. 

During inference, we are only required to perform the forward pass associated with $f_{\theta}(x_i)$ and the nearest neighbor lookup in the PL index, as the PL index itself can be constructed offline. This makes the approach very suitable for practical scenarios where the number of candidate code snippets $|\mathcal{C}|$ could be very large.

In a related line of work, \cite{lu2021codexglue} propose a benchmark (NL-code-search-WebQuery) where natural language code search is framed as the problem of analysing a query-code pair to predict whether the code answers the query or not. \cite{huang2021cosqa} release a new dataset with manually written queries (as opposed to docstrings extracted automatically), and propose a similar benchmark based on binary classification of query-code pairs.   
\section{CasCode}
\label{sec_method}

Although the approach proposed by \cite{guo2020graphcodebert} is efficient for practical scenarios, the independent encodings of the query and the code make it less effective. We could instead encode the query and the code candidate jointly within a single transformer encoder and perform binary classification. In particular, the model could take as input the concatenation of NL and PL sequences $[x_i ; y_j]$ and predict whether the two match in semantics.

\begin{wrapfigure}{r}{0.45\textwidth}
\vspace{-2em}
  \begin{center}
    \includegraphics[width=0.4\textwidth]{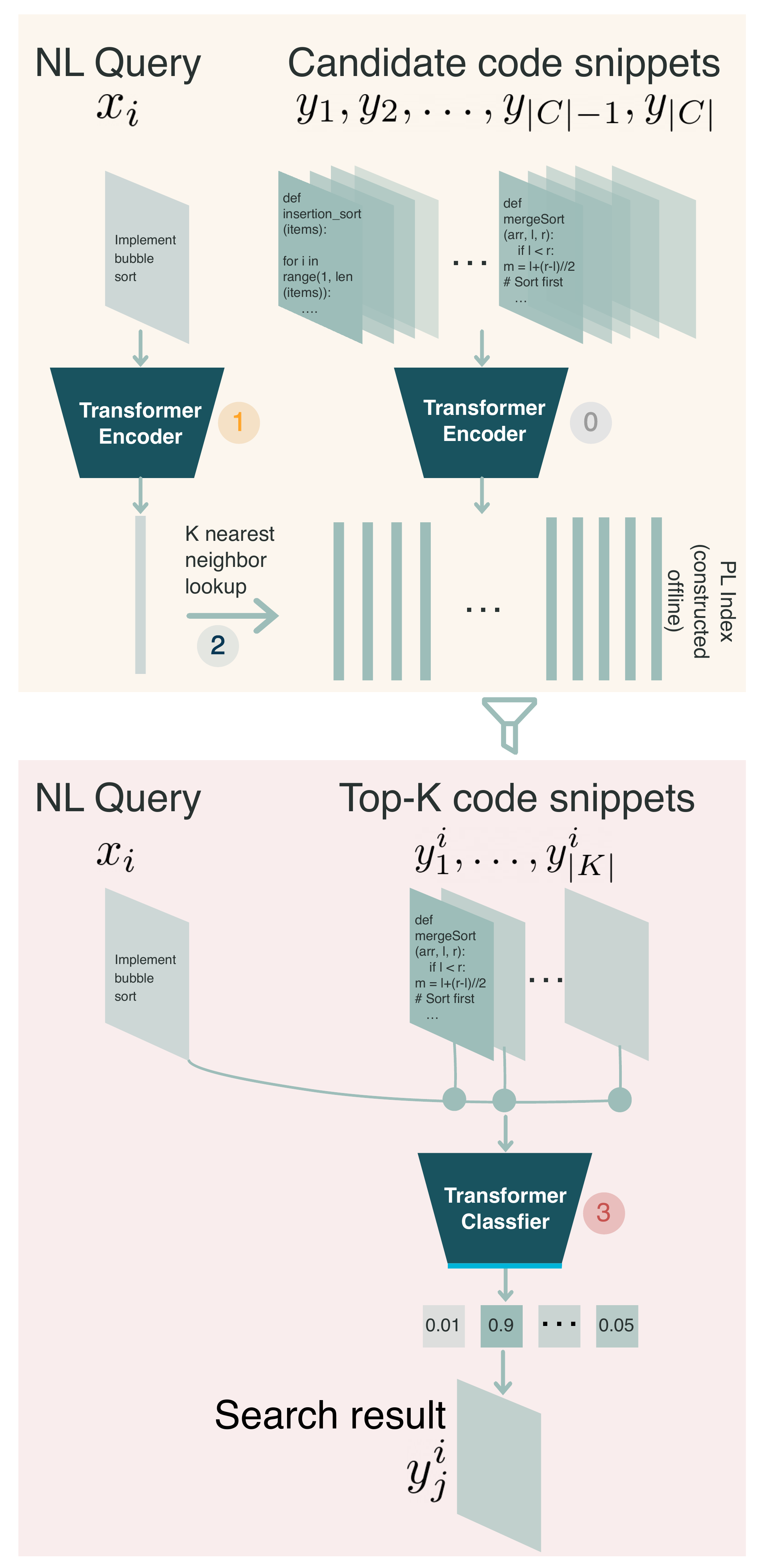}
  \end{center}
  \caption{CasCode: Our proposed cascaded scheme for semantic code search. At the top, the transformer encoder independently processes the query $x_i$ and the code snippets in the \textit{fast} retrieval stage. The top K candidates (based on the nearest neighbor lookup) from this stage are passed on to the second stage, where a transformer classifier jointly processes the query sequence with each of the filtered candidates to predict the probability of their semantics matching. The second stage classifiers are thus accelerated for the code retrieval task by the first stage of encoders.
  }
 \label{fig_cascode_main}
 \vspace{-4em}
\end{wrapfigure}

The training batches for this binary classification setup can again be constructed using the bimodal dataset (positive pairs denoting semantic matches), and the negative pairs (mismatch) can be constructed artificially. 
Given a set of paired NL-PL semantically equivalent sequences $\{x_i, y_i \}_{i=1}^{N}$, the cross-entropy objective function for this training scheme would be:
\begin{align}
\small 
\label{eqn:bin}
\mathcal{L}_{\text{CE}} = -\frac{1}{N} \hspace{-0.5em} \sum_{i=1, j \ne i}^{N} \hspace{-0.5em} \log p_{\theta}{(x_i,y_i)} + \log (1 - p_{\theta}{(x_i,y_j)})
\end{align}
where $p_{\theta}{(x_i,y_j)}$ represents the probability that the NL sequence $x_i$ semantically matches the PL sequence $y_j$, as predicted by the classifier. With a minibatch of positive pairs $\{x_i, y_i \} ~\forall i \in \mathcal{B}$, we can randomly pick $y_j$ ($j \in \mathcal{B}; j \neq i$) from the PL sequences in the minibatch and pair it with $x_i$ to serve as a negative pair. When using a transformer encoder based classifier, the interactions between the NL and PL tokens in the self-attention layers can help in improving the precision of this approach over the previous (independent encoding) one.

During inference, we can pair the NL sequence $x_i$ with each of the $y_j$ from $\mathcal{C}$ and rank the candidates as per the classifier's confidence scores of the pair being a match. This involves $\mathcal{C}$ forward passes (each on a joint NL-PL sequence, thus longer inputs than the previous approach), making this approach infeasible when dealing with large retrieval sets. We refer to this approach as the one using \textit{slow classifiers} for retrieval.  Figure \ref{fig_fast_slow} provides an illustration of these two different approaches.

We propose unifying the strengths of the two approaches - the speed of the \textit{fast encoders} with the precision of the \textit{slow classifiers}, with a cascaded scheme, called CasCode. Figure \ref{fig_cascode_main} shows the overall framework of our approach. Our hybrid strategy combines the strengths of the two approaches in the following manner - the first stage of \textit{fast encoders} provides top-$K$ candidates from the set $\mathcal{C}$ of candidate code snippets. In practice, the size of the retrieval set ($|\mathcal{C}|$) can often be very large, and varies from $4360$ to $52660$ for the CodeSearchNet datasets we study in our experiments. 

The top $K$ candidates are then passed to the second stage of \textit{slow classifiers} where each of them is paired with the NL input (query) $x_i$ and fed to the model. For a given pair, this second stage classifier will return the probability of the NL and PL components of the input matching in semantics. Using these as confidence scores, the rankings of the $K$ candidates are refined.

The resulting scheme is preferable for $K << |\mathcal{C}|$, as this would add a minor computational overhead on top of what is required by the \textit{fast encoder} based retrieval. The second stage of refinement can then improve retrieval performance provided that the value of $K$ is set such that the recall of the \textit{fast encoder} is reasonably high. $K$ would be a critical hyper-parameter in this scheme, as setting a very low $K$ would lead to high likelihood of missing the correct snippet in the set of inputs passed to the second stage \textit{slow classifier}, while a very high $K$ would make the scheme infeasible for retrieval. As we discuss ahead in Section \ref{sec_experiments}, CasCode with a $K$ as small as $10$ already offers significant gains in retrieval performance over the baselines, with marginal gains as we increment $K$ to $100$ and beyond.

In order to minimize the memory overhead incurred by the two stage model, we propose to share the weights of the transformer layers of the \textit{fast encoders} and the \textit{slow classifiers}. This can be achieved by training a model with the joint objective of infoNCE ($\mathcal{L}_{\text{infoNCE}}$) and binary cross-entropy ($\mathcal{L}_{\text{CE}}$.) While the number of parameters in this shared variant would be nearly half of the separat (non-shared) case, the computational cost at inference would be the same. Note that we would need some exclusive parameters for the classifier model, specifically the classification head (MLP) on top of the encoder. Thus, in this shared parameter variant of CasCode, the transformer model consuming the three kinds of inputs - NL only and PL only (for the \textit{fast encoder} stage) and NL-PL (for the \textit{slow classifier} stage) is identical except for the MLP layers in the second stage.

\section{Experiments}
\label{sec_experiments}

\subsection{Dataset, Baselines \& Metrics}

We use the CodeSearchNet code corpus from \cite{husain2019codesearchnet} that includes six programming languages - Ruby, Javascript, Go, Python, Java and Php. Our pre-processing and train-val-test splits are identical to the setting from \cite{guo2020graphcodebert}\footnote{https://github.com/microsoft/CodeBERT/tree/master/GraphCodeBERT}, who filter low-quality queries and expand the retrieval set to make the code search task more challenging and realistic. Table \ref{tab:cdsn_py_example} shows 2 examples of bimodal pairs from the resulting dataset and the statistics of the dataset after pre-processing are provided in Table \ref{codesearchnet_details}. 

\begin{table}[]
\centering
\begin{tabular}{ p{3.5cm} p{8.5cm}}
\toprule
\small
\texttt{Prompt the user to continue or not} & 
\vspace{-4mm}
\scriptsize
\begin{lstlisting}[language=Python]
def continue_prompt(message = ""):
    answer = False
    message = message + """\n"Yes" or "No" to continue: """
    
    while answer not in ("Yes", "No"):
        answer = prompt ( message, eventloop = eventloop())
        if answer == "Yes":
            break
        if answer == "No":
            break
	return answer
\end{lstlisting}
         \\
\vspace{-4mm}
\small
\texttt{Sends a message to the framework scheduler.} & 
\vspace{-4mm}
\scriptsize
\begin{lstlisting}[language=Python]
def message(self, data): 
	logging.info("""Driver sends framework 
	message {}""".format(data)) 
	return self.driver.sendFrameworkMessage(
	data)
\end{lstlisting}\\
\bottomrule
    \end{tabular}
    \caption{Examples of bimodal pairs (natural language/docstring with corresponding code sequence) from CodeSearchNet (Python)}
    \label{tab:cdsn_py_example}
\end{table}

\begin{table}
\centering
\begin{adjustbox}{max width=0.95\textwidth}
\begin{tabular}{l c c c c c c c}
\toprule
- & \textbf{Go} & \textbf{Java} & \textbf{Javascript} & \textbf{PHP} & \textbf{Python} & \textbf{Ruby}\\
\midrule
Training examples & 167,288 & 164,923 & 58,025 &  241,241 & 251,820 & 24,927 \\
Dev queries & 7,325 & 5,183 & 3,885 & 12,982 & 13,914 & 1,400\\
Testing queries & 8,122 & 10,955 & 3,291 &  14,014 & 14,918 & 1,261 \\
Candidate codes & 28,120 & 40,347 & 13,981 &  52,660 & 43,827 & 4,360 \\
\bottomrule
\end{tabular}
\end{adjustbox}
\caption{\label{citation-guide}
\small Data statistics of the filtered CodeSearchNet corpus for Go, Java, Javascript, PHP, Python and Ruby programming languages. For each query in the dev and test sets, the answer is retrieved from the set of candidate codes (last row)
}\label{codesearchnet_details}
\end{table}

\textbf{Our \textit{fast encoder} baseline} is based on the CodeBERT model from \cite{feng2020codebert} that is pre-trained on programming languages. In order to have a strong baseline, we use a newer CodeBERT checkpoint that is pre-trained (using masked language modeling and replaced token detection tasks) for longer, after we found that the CodeBERT checkpoint from \cite{feng2020codebert} was not trained till convergence. When starting from our new checkpoint, we find that the CodeBERT baseline, if fine-tuned with a larger batch-size (largest possible that we can fit on 8 A100 GPUs) and for a larger number of epochs, is able to perform substantially better than the results reported before. We report the baselines from \cite{guo2020graphcodebert} in Table \ref{tab:main-table} along with the results for our replication of two of these baselines. Previous studies have emphasized this effect - larger batch sizes are known to typically work well when training with the infoNCE loss in a contrastive learning framework, due to more negative samples from the batch~\citep{chen2020simple}.

We also train GraphCodeBERT, which is proposed by \cite{guo2020graphcodebert} as a structure aware model pre-trained on programming languages. GraphCodeBERT leverages data flow graphs during pre-training to incorporate structural information into its representations. However, for the code search task, we report (Table \ref{tab:main-table}) that GraphCodeBERT does not offer any significant improvements in performance over CodeBERT, when both variants are trained with a large batch size. For simplicity, we finetune the CodeBERT pre-trained model (architecturally equivalent to RoBERTa-base\cite{liu2019roberta} model - 12 layers, 768 dimensional hidden states and 12 attention heads) and refer this as the \textit{fast encoder} baseline for the remainder of our experiments.

\begin{table}[t]
\centering
\begin{adjustbox}{max width=1.0\textwidth}
\begin{tabular}{p{3cm} c c c c c c c}
\toprule
\textbf{Model/Method} & \textbf{Ruby} & \textbf{Javascript} & \textbf{Go} & \textbf{Python} & \textbf{Java} & \textbf{Php} & \textbf{Overall}\\
\midrule
NBow & 0.162 & 0.157 & 0.330 & 0.161 & 0.171 & 0.152 & 0.189\\
CNN & 0.276 & 0.224 & 0.680 & 0.242 & 0.263 & 0.260 & 0.324\\
BiRNN & 0.213 & 0.193 & 0.688 & 0.290 & 0.304 & 0.338 & 0.338\\
selfAtt & 0.275 & 0.287 & 0.723 & 0.398 & 0.404 & 0.426 & 0.419\\
\midrule
\multicolumn{8}{c}{\textit{As reported by \cite{guo2020graphcodebert}}} \\[+0.5em]
RoBERTa & 0.587 & 0.517 & 0.850 & 0.587 & 0.599 & 0.560 & 0.617\\
RoBERTa (code) & 0.628 & 0.562 & 0.859 & 0.610 & 0.620 & 0.579 & 0.643\\
CodeBERT & 0.679 & 0.620 & 0.882 & 0.672 & 0.676 & 0.618 & 0.693\\
GraphCodeBERT &  0.703 & 0.644 & 0.897 & 0.692 & 0.691 & 0.649 & 0.713\\
\midrule
\multicolumn{8}{c}{\textit{As reported by \cite{wang2022syncobert} }} \\[+0.5em]
SYNCOBERT  & 0.722 & 0.677 & 0.913 & 0.724 & 0.723 & 0.678 & 0.740 \\ 
\midrule
\multicolumn{8}{c}{\textit{Replicated with a larger training batch-size}} \\[+0.5em]
CodeBERT & 0.7245 & 0.6794 & 0.9145 & 0.7305 & 0.7317 & 0.681 & 0.7436\\
GraphCodeBERT & 0.7253 & 0.6722 & 0.9157 & 0.7288 & 0.7275 & 0.6835 & 0.7422 \\
\midrule
\multicolumn{8}{c}{\textit{Ours ($K$=10)}}  \\ [+0.5em]
CasCode (shared) & 0.7621 & 0.6948 & 0.9193 & 0.7529 & 0.7528 & 0.7001 & 0.7637 \\
CasCode (separate) & 0.7724 & 0.7087 & 0.9258 & 0.7645 & 0.7623 & 0.7028 & 0.7727 \\
\midrule
\multicolumn{8}{c}{\textit{Ours ($K$=100)}}  \\ [+0.5em]
CasCode (shared) & 0.7686 & 0.6989 & 0.9232 & 0.7618 & 0.7602 & 0.7074 & 0.77 \\
CasCode (separate) & \textbf{0.7825} & \textbf{0.716} & \textbf{0.9272} & \textbf{0.7704} & \textbf{0.7723} & \textbf{0.7083} & \textbf{0.7795} \\
\bottomrule
\end{tabular}
\end{adjustbox}
\caption{
Mean Reciprocal Ranking (MRR) values of different methods on the codesearch task on 6 Programming Languages from the CodeSearchNet corpus (test set). The first set consists of four finetuning-based baseline methods (NBow: Bag of words, CNN: convolutional neural network, BiRNN: bidirectional recurrent neural network, and multi-head attention), followed by the second set of models that are pre-trained then finetuned for code search (RoBERTa: pre-trained on text by  \citet{liu2019roberta}, RoBERTa (code): RoBERTa pre-trained only on code, CodeBERT: pre-trained on code-text pairs by \citet{feng-etal-2020-codebert}, GraphCodeBERT: pre-trained using structure-aware tasks by \citet{guo2020graphcodebert}). SYNCOBERT: pre-trained using syntax-aware tasks by \citet{wang2022syncobert}. In the last four rows, we report the results with the shared and separate variants of our CasCode scheme using the fine-tuned CodeBERT models for $K$ of $10$ and $100$.
}\label{tab:main-table}
\end{table}

MRR score for this CodeBERT baseline is shown in Table \ref{tab:main-table}. For this baseline and the variants we propose, along with MRR, we also report Recall@K for $K = \{1,2,5,8,10\}$, that indicates the hit rate (ratio of instances where we find the correct output in the top $K$ results). We encourage future work on code search to report these additional metrics, as these are important in evaluating the utility of a retrieval system and is commonly reported in similar work in text based image or video retrieval~\citep{miech2021thinking,frozen2021}.

Figure \ref{fig_recall_fast} shows the Recall@K (K varied over the horizontal axis) for the 6 different programming languages, with the \textit{fast encoder} models, over the validation set. As alluded to in Section \ref{sec_method}, for designing the cascaded scheme, we need to pick a $K$ that is large enough to provide reasonably high recall, and small enough for the second stage to be reasonably fast. We pick $K = 10 \text{ and } 100$ where the recall for all 6 datasets is over $85\% \text{ and } 90 \%$ respectively.

\subsection{Results with CasCode}

\begin{figure}
    \centering
    \begin{minipage}{0.48\textwidth}
        \centering
        \includegraphics[width=0.97\textwidth]{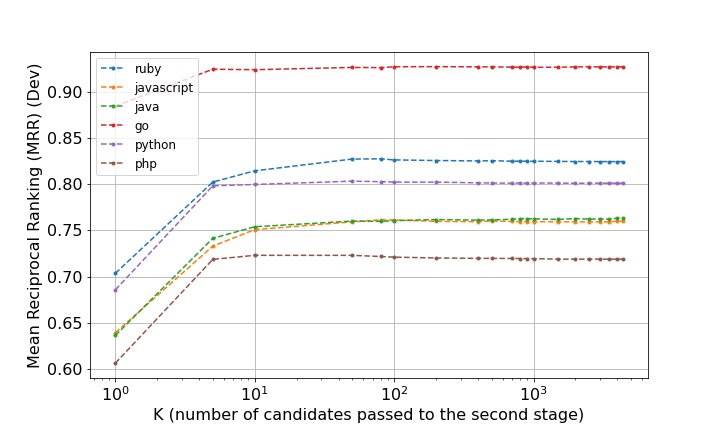} 
        \caption{\small Mean reciprocal ranking (MRR) at different values of K over the validation set of CodeSearchNet \cite{husain2019codesearchnet} when using a finetuned CodeBERT (\textit{slow}) binary classifier (match or not) for text-code retrieval.}
        \label{fig_mrr_slow}
    \end{minipage}\hfill
    \begin{minipage}{0.48\textwidth}
        \centering
        \includegraphics[width=1\textwidth]{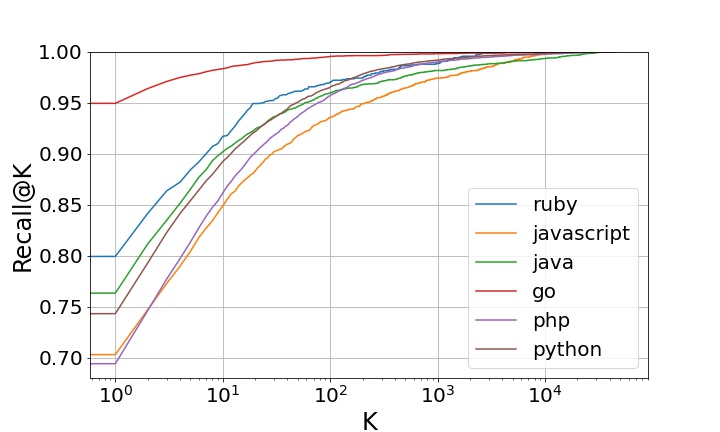} 
        \caption{\small Recall at different values of K over the validation set of CodeSearchNet \cite{husain2019codesearchnet} when using a finetuned CodeBERT encoder (\textit{fast}) for text-code retrieval.}
        \label{fig_recall_fast}
    \end{minipage}
\end{figure}

We first show results for the \textbf{\textit{slow classifier}}s, trained using the CodeSearhNet datasets that we mention above. We finetune the CodeBERT pre-trained checkpoint (mentioned above) with a classification head (fully connected layers) for this task. On the validation set, we study the performance of this finetuned classifier for retrieval and report the MRR scores in Figure \ref{fig_mrr_slow} for different values of $K$, where $K$ is the number of top candidates passed from the first (\textit{fast encoder}) stage to the second. Interestingly, the retrieval performance of this joint classifier fails to improve beyond certain values of $K$. For example, increasing $K$ from $10$ to $100$ only marginally improves the MRR for Ruby, Javascript and Java, while for other languages there is no significant improvement beyond $K = 10$. Further training details for CasCode variants and the \textit{fast encoder} baselines are provided in Appendix \ref{appendix}. 

Next, we train \textit{fast} and \textit{slow} models with \textbf{shared parameters}, denoted by CasCode (shared). The training objective for this model is the average of the binary cross-entropy loss $\mathcal{L}_{\text{CE}}$ and the infoNCE loss $\mathcal{L}_{\text{infoNCE}}$ as described in Section \ref{sec_method}. The MRR scores for the baselines and our separate and shared variants are listed in Table \ref{tab:main-table}. With our cascaded approach, we observe significant improvements over the \textit{fast encoder} baselines, the overall MRR\footnote{we report MRR on the scale of 0-1, some works (eg. \cite{wang2022syncobert}) use the scale 0-100} averaged over the six programming languges for CasCode (separate) is $0.7795$, whereas the \textit{fast encoder} baseline (CodeBERT) reaches $0.7422$. The improvements with CasCode are noticeably greater over the baseline for Ruby, Javascript, Python and Java. We report modest improvements on the Go dataset, where the \textit{fast encoder} baseline is already quite strong ($0.9145$ MRR).

The shared variant of CasCode attains an overall MRR score of $0.77$, which is comparable to the separate variant performance. This slight difference can be attributed to the limited model capacity in the shared case, as the same set of transformer layers serve in the encoder and classifier models. We also evaluate the MRR scores for the CasCode (shared) model in the \textit{fast encoder} stage where the test set MRR scores were $0.7308, 0.6634, 0.9048, 0.7193, 0.7244, 0.6803$ for Ruby, Javascript, Go, Python, Java and PHP respectively, with the overall MRR being $0.7372$. We note in passing, that the cascaded model that was trained in a multi-task manner, gives competitive retrieval performance, even when used only in its first (encoder only) stage.

\begin{figure}[h]
  \begin{center}
    \includegraphics[scale=0.35]{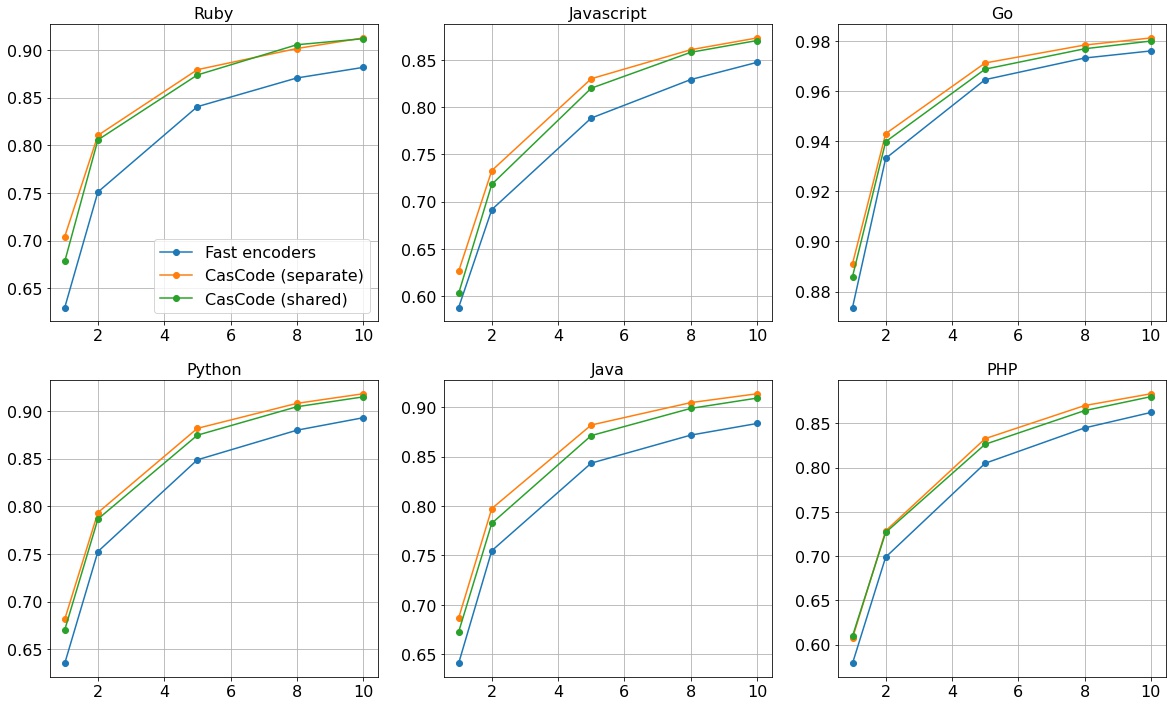}
  \end{center}
  \caption{Recall @ K = $\{1, 2, 5, 8, 10 \}$ with the \textit{fast encoder} and CasCode (shared and separate) methods on the test set queries of CodeSearchNet dataset.}
  \label{fig_recall_variants}
\end{figure}

We also report the Recall@K metric for CasCode separate and CasCode shared variants in Figure \ref{fig_recall_variants}. For all six programming languages, we observe improvements over the \textit{fast encoder} baseline with our cascaded scheme. Similar to our observation from Table \ref{tab:main-table}, the shared variant of CasCode is slightly worse than the separate one.

\paragraph{Retrieval speed comparison:} Having established the improvements in retrieval performance with CasCode, we proceed to analyze the trade-off between inference speed and performance, for the different methods discussed. For each variant, we record the time duration (averaged over $100$ instances) required to process (obtain a relevant code snippet from the retrieval set) a natural langugae query from the held-out set. We use the Ruby dataset of CodeSearchNet for this analysis, which contains $4360$ candidate code snippets for each NL query. We conduct this study on a single Nvidia A100 GPU. Our results are shown in Table \ref{speed_comparison}.

For the \textit{fast encoder} approach (using infoNCE-finetuned CodeBERT), we first incur some computational cost to encode all the candidate code snippets and construct the PL index ($6.76$ seconds for Ruby's retrieval set). This computation is common to all approaches, except the \textit{slow} (binary, joint) classifier one. Since this computation can be performed offline before the model is deployed to serve user queries, we do not include this cost in our results in Table \ref{speed_comparison}. With the PL index constructed beforehand, we report the time required to encode a user NL query, and perform nearest neighbor lookup on the PL index with the encoding, in the first row of Table \ref{speed_comparison}. This computation is again performed by all the CasCode variants, and thus acts as the lower bound on time taken by CasCode for retrieval. For the analysis to be as close to real world scenarios as possible, we do not batch the queries and encode them one by one. Batching them would require assuming that we have the NL queries beforehand, while we would be receiving them on the fly from users when deployed.

\begin{table}[t]
\centering
\begin{tabular}{c c c c c}\\\toprule  
Model & $\#$ params & Inference duration (secs) & MRR & $\#$ queries/s \\\midrule
Fast encoders (CodeBERT) & 125M & 0.0427  & 0.7245 & 23.42 \\  \midrule
Slow Binary classifiers & 125M + 0.5M & 9.1486 & 0.7816 & 0.11 \\  \midrule
CasCode (separate, K=100) & 250M + 0.5M & 0.2883 & 0.7825 & 3.46 \\  \midrule
CasCode (shared, K=100) & 125M + 0.5M & 0.2956 & 0.7686 & 3.38 \\ \midrule
CasCode (separate, K=10) & 250M + 0.5M & 0.1022 & 0.7724 & 9.78 \\  \midrule
CasCode (shared, K=10) & 125M + 0.5M & 0.1307 & 0.7621 & 7.65 \\ \midrule
\bottomrule
\end{tabular}
\caption{Inference speed comparison for the different variants studied. The number of parameters corresponding to the classifier head are separated with a $`+\textrm'$ sign in the second column. Inference duration is averaged for $100$ queries from the Ruby subset of CodeSearchNet, using a single A100 GPU. Constructing the PL index offline requires $6.76$ seconds for the Ruby dataset and is not included in the durations listed here. MRR scores are reported on the entire test set. Throughput of the retrieval model (measured in $\#$ queries processed per second) is listed in the last column. }\label{speed_comparison}
\end{table}

With the \textit{slow classifier} approach, we would pair a given query with each of the $4360$ candidates, and thus this would lead to the slowest inference of all the variants. For all variants of CasCode, the inference duration listed in Table \ref{speed_comparison} includes the time taken by the \textit{fast encoder} based retrieval (first stage). For CasCode's second stage, we can pass the $K$ combinations (query concatenated with each of the top-$K$ candidate from the fast stage) in a batched manner. The shared variant, while requiring half the parameters, incurs the same computational cost when used in the cascaded fashion. We note from Table \ref{speed_comparison} that at a minor drop in the MRR score, lowering CasCode's $K$ from $100$ can lead to almost $3$x faster inference for the shared case.

\section{Conclusion \& Future work}
\label{sec_conclusion}

We propose CasCode, which is a cascaded scheme consisting of transformer encoder and joint binary classifier stages for the task of semantic code search and achieve state of the art performance on the CodeSearchNet benchmark, with significant improvements over previous results. We also propose a shared parameter variant of CasCode, where a single transformer encoder can operate in the two different stages when trained in a multi-task fashion. At almost half the number of parameters, CasCode's shared variant offers comparable performance to the non-shared (separate) variant. 

A limitation of our current cascaded scheme is that the computation spent in generating representations in the first stage of \textit{fast encoders} is not leveraged in the second stage. We process raw token level inputs in the second stage. Ideally the representations designed in the first stage should be useful for the classification stage too \citep{ALBEF}. Our initial attempts along this direction did not turn fruitful, and future work could address this aspect. Another limitation warranting further investigation is associated with the training of the shared variant of CasCode. Here, training with the multitask learning framework (joint objective of infoNCE and binary cross entropy) leads to a model that performs slightly worse than the separate variant (individually finetuned models). We tried augmenting the capabilites of this model with solutions like using independent CLS tokens for the three modes the model has to operate in (NL only, PL only, NL-PL concatenation), and adjusting the relative weight of the two losses involved, but could not achieve any improvement over the separate variant.

\bibliography{iclr2022_conference}

\begin{thebibliography}{25}
\providecommand{\natexlab}[1]{#1}
\providecommand{\url}[1]{\texttt{#1}}
\expandafter\ifx\csname urlstyle\endcsname\relax
  \providecommand{\doi}[1]{doi: #1}\else
  \providecommand{\doi}{doi: \begingroup \urlstyle{rm}\Url}\fi

\bibitem[Ahmad et~al.(2021)Ahmad, Chakraborty, Ray, and
  Chang]{ahmad-etal-2021-unified}
Wasi Ahmad, Saikat Chakraborty, Baishakhi Ray, and Kai-Wei Chang.
\newblock Unified pre-training for program understanding and generation.
\newblock In \emph{Proceedings of the 2021 Conference of the North American
  Chapter of the Association for Computational Linguistics: Human Language
  Technologies}, pp.\  2655--2668, Online, June 2021. Association for
  Computational Linguistics.
\newblock \doi{10.18653/v1/2021.naacl-main.211}.
\newblock URL \url{https://aclanthology.org/2021.naacl-main.211}.

\bibitem[Austin et~al.(2021)Austin, Odena, Nye, Bosma, Michalewski, Dohan,
  Jiang, Cai, Terry, Le, et~al.]{austin2021program}
Jacob Austin, Augustus Odena, Maxwell Nye, Maarten Bosma, Henryk Michalewski,
  David Dohan, Ellen Jiang, Carrie Cai, Michael Terry, Quoc Le, et~al.
\newblock Program synthesis with large language models.
\newblock \emph{arXiv preprint arXiv:2108.07732}, 2021.

\bibitem[Bain et~al.(2021)Bain, Nagrani, Varol, and Zisserman]{frozen2021}
Max Bain, Arsha Nagrani, G{\"u}l Varol, and Andrew Zisserman.
\newblock Frozen in time: A joint video and image encoder for end-to-end
  retrieval.
\newblock In \emph{IEEE International Conference on Computer Vision}, 2021.

\bibitem[Cambronero et~al.(2019)Cambronero, Li, Kim, Sen, and
  Chandra]{cambronero2019deep}
Jose Cambronero, Hongyu Li, Seohyun Kim, Koushik Sen, and Satish Chandra.
\newblock When deep learning met code search.
\newblock In \emph{Proceedings of the 2019 27th ACM Joint Meeting on European
  Software Engineering Conference and Symposium on the Foundations of Software
  Engineering}, pp.\  964--974, 2019.

\bibitem[Chen et~al.(2021)Chen, Tworek, Jun, Yuan, Ponde, Kaplan, Edwards,
  Burda, Joseph, Brockman, et~al.]{chen2021evaluating}
Mark Chen, Jerry Tworek, Heewoo Jun, Qiming Yuan, Henrique Ponde, Jared Kaplan,
  Harri Edwards, Yura Burda, Nicholas Joseph, Greg Brockman, et~al.
\newblock Evaluating large language models trained on code.
\newblock \emph{arXiv preprint arXiv:2107.03374}, 2021.

\bibitem[Chen et~al.(2020)Chen, Kornblith, Norouzi, and Hinton]{chen2020simple}
Ting Chen, Simon Kornblith, Mohammad Norouzi, and Geoffrey Hinton.
\newblock A simple framework for contrastive learning of visual
  representations.
\newblock In \emph{International conference on machine learning}, pp.\
  1597--1607. PMLR, 2020.

\bibitem[Devlin et~al.(2019)Devlin, Chang, Lee, and
  Toutanova]{devlin-etal-2019-bert}
Jacob Devlin, Ming-Wei Chang, Kenton Lee, and Kristina Toutanova.
\newblock {BERT}: Pre-training of deep bidirectional transformers for language
  understanding.
\newblock In \emph{Proceedings of the 2019 Conference of the North {A}merican
  Chapter of the Association for Computational Linguistics: Human Language
  Technologies, Volume 1 (Long and Short Papers)}, pp.\  4171--4186,
  Minneapolis, Minnesota, June 2019. Association for Computational Linguistics.
\newblock \doi{10.18653/v1/N19-1423}.
\newblock URL \url{https://aclanthology.org/N19-1423}.

\bibitem[Feng et~al.(2020{\natexlab{a}})Feng, Guo, Tang, Duan, Feng, Gong,
  Shou, Qin, Liu, Jiang, and Zhou]{feng-etal-2020-codebert}
Zhangyin Feng, Daya Guo, Duyu Tang, Nan Duan, Xiaocheng Feng, Ming Gong, Linjun
  Shou, Bing Qin, Ting Liu, Daxin Jiang, and Ming Zhou.
\newblock {C}ode{BERT}: A pre-trained model for programming and natural
  languages.
\newblock In \emph{Findings of the Association for Computational Linguistics:
  EMNLP 2020}, pp.\  1536--1547, Online, November 2020{\natexlab{a}}.
  Association for Computational Linguistics.
\newblock \doi{10.18653/v1/2020.findings-emnlp.139}.
\newblock URL \url{https://www.aclweb.org/anthology/2020.findings-emnlp.139}.

\bibitem[Feng et~al.(2020{\natexlab{b}})Feng, Guo, Tang, Duan, Feng, Gong,
  Shou, Qin, Liu, Jiang, et~al.]{feng2020codebert}
Zhangyin Feng, Daya Guo, Duyu Tang, Nan Duan, Xiaocheng Feng, Ming Gong, Linjun
  Shou, Bing Qin, Ting Liu, Daxin Jiang, et~al.
\newblock Codebert: A pre-trained model for programming and natural languages.
\newblock \emph{arXiv preprint arXiv:2002.08155}, 2020{\natexlab{b}}.

\bibitem[Gu et~al.(2018)Gu, Zhang, and Kim]{gu2018deep}
Xiaodong Gu, Hongyu Zhang, and Sunghun Kim.
\newblock Deep code search.
\newblock In \emph{2018 IEEE/ACM 40th International Conference on Software
  Engineering (ICSE)}, pp.\  933--944. IEEE, 2018.

\bibitem[Guo et~al.(2021)Guo, Ren, Lu, Feng, Tang, Liu, Zhou, Duan,
  Svyatkovskiy, Fu, et~al.]{guo2020graphcodebert}
Daya Guo, Shuo Ren, Shuai Lu, Zhangyin Feng, Duyu Tang, Shujie Liu, Long Zhou,
  Nan Duan, Alexey Svyatkovskiy, Shengyu Fu, et~al.
\newblock Graphcodebert: Pre-training code representations with data flow.
\newblock \emph{ICLR 2021}, 2021.

\bibitem[Gutmann \& Hyv{\"a}rinen(2010)Gutmann and
  Hyv{\"a}rinen]{gutmann2010noise}
Michael Gutmann and Aapo Hyv{\"a}rinen.
\newblock Noise-contrastive estimation: A new estimation principle for
  unnormalized statistical models.
\newblock In \emph{Proceedings of the thirteenth international conference on
  artificial intelligence and statistics}, pp.\  297--304. JMLR Workshop and
  Conference Proceedings, 2010.

\bibitem[Huang et~al.(2021)Huang, Tang, Shou, Gong, Xu, Jiang, Zhou, and
  Duan]{huang2021cosqa}
Junjie Huang, Duyu Tang, Linjun Shou, Ming Gong, Ke~Xu, Daxin Jiang, Ming Zhou,
  and Nan Duan.
\newblock Cosqa: 20,000+ web queries for code search and question answering.
\newblock \emph{arXiv preprint arXiv:2105.13239}, 2021.

\bibitem[Husain et~al.(2019)Husain, Wu, Gazit, Allamanis, and
  Brockschmidt]{husain2019codesearchnet}
Hamel Husain, Ho-Hsiang Wu, Tiferet Gazit, Miltiadis Allamanis, and Marc
  Brockschmidt.
\newblock Codesearchnet challenge: Evaluating the state of semantic code
  search.
\newblock \emph{arXiv preprint arXiv:1909.09436}, 2019.

\bibitem[Li et~al.(2021)Li, Selvaraju, Gotmare, Joty, Xiong, and Hoi]{ALBEF}
Junnan Li, Ramprasaath~R. Selvaraju, Akhilesh~Deepak Gotmare, Shafiq Joty,
  Caiming Xiong, and Steven Hoi.
\newblock Align before fuse: Vision and language representation learning with
  momentum distillation.
\newblock In \emph{NeurIPS}, 2021.

\bibitem[Liu et~al.(2019)Liu, Ott, Goyal, Du, Joshi, Chen, Levy, Lewis,
  Zettlemoyer, and Stoyanov]{liu2019roberta}
Yinhan Liu, Myle Ott, Naman Goyal, Jingfei Du, Mandar Joshi, Danqi Chen, Omer
  Levy, Mike Lewis, Luke Zettlemoyer, and Veselin Stoyanov.
\newblock Roberta: A robustly optimized bert pretraining approach.
\newblock \emph{arXiv preprint arXiv:1907.11692}, 2019.

\bibitem[Lu et~al.(2021)Lu, Guo, Ren, Huang, Svyatkovskiy, Blanco, Clement,
  Drain, Jiang, Tang, et~al.]{lu2021codexglue}
Shuai Lu, Daya Guo, Shuo Ren, Junjie Huang, Alexey Svyatkovskiy, Ambrosio
  Blanco, Colin Clement, Dawn Drain, Daxin Jiang, Duyu Tang, et~al.
\newblock Codexglue: A machine learning benchmark dataset for code
  understanding and generation.
\newblock \emph{arXiv preprint arXiv:2102.04664}, 2021.

\bibitem[Miech et~al.(2021)Miech, Alayrac, Laptev, Sivic, and
  Zisserman]{miech2021thinking}
Antoine Miech, Jean-Baptiste Alayrac, Ivan Laptev, Josef Sivic, and Andrew
  Zisserman.
\newblock Thinking fast and slow: Efficient text-to-visual retrieval with
  transformers.
\newblock In \emph{Proceedings of the IEEE/CVF Conference on Computer Vision
  and Pattern Recognition}, pp.\  9826--9836, 2021.

\bibitem[Sachdev et~al.(2018)Sachdev, Li, Luan, Kim, Sen, and
  Chandra]{sachdev2018retrieval}
Saksham Sachdev, Hongyu Li, Sifei Luan, Seohyun Kim, Koushik Sen, and Satish
  Chandra.
\newblock Retrieval on source code: a neural code search.
\newblock In \emph{Proceedings of the 2nd ACM SIGPLAN International Workshop on
  Machine Learning and Programming Languages}, pp.\  31--41, 2018.

\bibitem[Svyatkovskiy et~al.(2020)Svyatkovskiy, Deng, Fu, and
  Sundaresan]{svyatkovskiy2020intellicode}
Alexey Svyatkovskiy, Shao~Kun Deng, Shengyu Fu, and Neel Sundaresan.
\newblock Intellicode compose: Code generation using transformer.
\newblock In \emph{Proceedings of the 28th ACM Joint Meeting on European
  Software Engineering Conference and Symposium on the Foundations of Software
  Engineering}, pp.\  1433--1443, 2020.

\bibitem[Vaswani et~al.(2017)Vaswani, Shazeer, Parmar, Uszkoreit, Jones, Gomez,
  Kaiser, and Polosukhin]{VaswaniSPUJGKP17}
Ashish Vaswani, Noam Shazeer, Niki Parmar, Jakob Uszkoreit, Llion Jones,
  Aidan~N. Gomez, Lukasz Kaiser, and Illia Polosukhin.
\newblock Attention is all you need.
\newblock In Isabelle Guyon, Ulrike von Luxburg, Samy Bengio, Hanna~M. Wallach,
  Rob Fergus, S.~V.~N. Vishwanathan, and Roman Garnett (eds.), \emph{Advances
  in Neural Information Processing Systems 30: Annual Conference on Neural
  Information Processing Systems 2017, December 4-9, 2017, Long Beach, CA,
  {USA}}, pp.\  5998--6008, 2017.
\newblock URL
  \url{https://proceedings.neurips.cc/paper/2017/hash/3f5ee243547dee91fbd053c1c4a845aa-Abstract.html}.

\bibitem[Wang et~al.(2021{\natexlab{a}})Wang, Yasheng~Wang, Zhou, Wan, Liu, Li,
  Wu, Liu, and Jiang]{wang2022syncobert}
Xin Wang, Fei~Mi Yasheng~Wang, Pingyi Zhou, Yao Wan, Xiao Liu, Li~Li, Hao Wu,
  Jin Liu, and Xin Jiang.
\newblock Syncobert: Syntax-guided multi-modal contrastive pre-training for
  code representation.
\newblock \emph{arXiv preprint arXiv:2108.04556}, 2021{\natexlab{a}}.

\bibitem[Wang et~al.(2021{\natexlab{b}})Wang, Wang, Joty, and
  Hoi]{wang2021codet5}
Yue Wang, Weishi Wang, Shafiq Joty, and Steven~CH Hoi.
\newblock Codet5: Identifier-aware unified pre-trained encoder-decoder models
  for code understanding and generation.
\newblock \emph{Proceedings of the 2021 Conference on Empirical Methods in
  Natural Language Processing, EMNLP 2021}, 2021{\natexlab{b}}.

\bibitem[Xu et~al.(2021)Xu, Vasilescu, and Neubig]{xu2021ide}
Frank~F Xu, Bogdan Vasilescu, and Graham Neubig.
\newblock In-ide code generation from natural language: Promise and challenges.
\newblock \emph{arXiv preprint arXiv:2101.11149}, 2021.

\bibitem[Ye et~al.(2016)Ye, Shen, Ma, Bunescu, and Liu]{ye2016word}
Xin Ye, Hui Shen, Xiao Ma, Razvan Bunescu, and Chang Liu.
\newblock From word embeddings to document similarities for improved
  information retrieval in software engineering.
\newblock In \emph{Proceedings of the 38th international conference on software
  engineering}, pp.\  404--415, 2016.

\end{thebibliography}
\bibliographystyle{iclr2022_conference}

\clearpage

\appendix

\section{appendix}  
\label{appendix}
\paragraph{Training details:} We begin with the baseline implementation of GraphCodeBERT (publicly available) and adapt their codebase to also implement the CodeBERT model. For the cascaded schemes, many of our training design decisions are therefore the same as GraphCodeBERT. 

We use 8 A100 GPUs (each with $40$ GB RAM) to train our baselines and CasCode variants. During training, we set the batch-size to a value that occupies as much available GPU RAM as possible. This happens to be 576 for the CodeBERT and GraphCodeBERT baseline finetuning with the infoNCE loss (fast encoders). For training the joint NL-PL classifier of CasCode (separate), we use a batch size of 216. For CasCode (shared), we need to further reduce the batch size to 160. All models are trained for 100 epochs. 

For all our experiments we use a learning rate of 2e-5. We use the Adam optimizer to update model parameters and perform early stopping on the development set. For the CasCode variants, when performing evaluation on the development set, we use $K = 100$ candidates from the fast encoder stage. Using 8 A100 GPUs, rough training durations for CasCode on the Ruby, Javascript, Go, Python, Java and PHP datasets are 6.5, 8, 33, 41, 15 and 21 hours respectively with separate variant and 13, 17, 38, 42, 55.5, 56 hours with the sharef variant. We typically perform evaluation on the validation set once every epoch, but make it infrequent in some cases to speed up training for larger datasets like Python and PHP. Given the significant amount of computation invested in training these retrieval models, we plan to release these checkpoints to avoid wasteful redundant training and encourage future work on semantic code search.

\end{document}